%% file: deep-learning.tex
\documentclass[12pt]{article} 
\usepackage[slantedGreek]{mathpazo}
\usepackage{amsmath,amssymb,amsthm,amsfonts}
\usepackage[left=1.25in,top=1.0in,right=1.25in,bottom=1.0in]{geometry}
\usepackage[pdftex]{graphicx}
\usepackage{mathtools}
\usepackage{subcaption}
\usepackage{suffix}
\usepackage{color}
\usepackage{todonotes}
\usepackage{setspace}

\usepackage{datetime}

\RequirePackage[hyphens]{url}
\RequirePackage[colorlinks,citecolor=blue,urlcolor=blue]{hyperref}
\usepackage[authoryear]{natbib}

\newdateformat{monthyear}{\monthname[\THEMONTH] \THEYEAR}

\allowdisplaybreaks

\include{math-commands}

\graphicspath{{../../figures/}{../figures/}{./figures/}{./}}

\vspace{-0.7in}
\title{A Statistical Theory of Deep Learning via Proximal Splitting}
\author{
  Nicholas G. Polson \\
  \textit{Booth School of Business} \\
  \textit{University of Chicago} 
  \footnote{\scriptsize Professor of Econometrics and Statistics 
    at the Chicago Booth School of Business.
    ngp@chicagobooth.edu.
  }
  \and
  Brandon T. Willard \\
  \textit{Booth School of Business} \\
  \textit{University of Chicago} 
  \footnote{bwillard@uchicago.edu } 
  \and
  Massoud Heidari \\
  \textit{Sutton Place} \\
  \textit{New York} 
  \footnote{ email: heidari.massoud@gmail.com}
}
\date{First Draft: May 2015\\
This Draft: \monthyear\today{}}

\begin{document}

\maketitle
\begin{abstract}
  \noindent 
  In this paper we develop a statistical theory and an implementation of deep
  learning (DL) models. We show that an  elegant variable splitting scheme for
  the alternating direction method of multipliers (ADMM) optimises a deep
  learning objective.  We allow for non-smooth non-convex regularisation
  penalties to induce sparsity in parameter weights.  We provide a link between
  traditional shallow layer statistical models such as principal component and
  sliced inverse regression and deep layer models. We also define the degrees
  of freedom of a deep learning predictor and a predictive MSE criteria to
  perform model selection for comparing architecture designs. We focus on deep
  multi-class logistic learning although our methods apply more generally.  Our
  results suggest an interesting and previously under-exploited relationship
  between deep learning and proximal splitting techniques.  To illustrate our
  methodology, we provide a multi-class logit classification analysis of Fisher's Iris data  where we illustrate
  the convergence of our algorithm.  Finally, we conclude
  with directions for future research.  
  
  \vspace{0.1in}
  \noindent Keywords: Deep Learning, Sparsity, Dropout, Convolutional Neural Nets; 
  Regularisation; Bayesian MAP; Image Segmentation; Classification; Multi-class Logistic regression.
\end{abstract}

\newpage
\singlespacing
\section{Introduction}

Deep Learning (DL) provides a powerful tool for high dimensional data
reduction.  Many areas of applications in predictive modeling occur in
artificial intelligence and machine learning; including pattern recognition
\citet{ripley_pattern_1996}; computer vision \citet{dean_large_2012}; image
segmentation and scene parsing \citet{farabet_learning_2013}; predictive
diagnostics; intelligent gaming \citet{mnih_playing_2013}.  The salient feature
of a deep learning model is a predictive rule comprised by a layered
composition of link or activation functions. Deep architectures with at least
three layers have been shown to provide improved predictive performance
compared to traditional shallow architectures in a number of applications.  The
challenge for deep learning methodologies, however, are computational: the
objective function which measures model fit is typically highly multi-modal and
hard to optimise efficiently.  

We build on the extensive deep learning literature by showing that proximal
Bayesian optimisation techniques provide a
turn-key solution to estimation and optimisation of such models and for
calculating a regularisation path.  We allow for the possibility of
irregular non-convex regularisation penalties to induce sparsity in the deep
layer weights.  Proximal operators and the alternating direction method of
multipliers (ADMM) are the key tools for implementation. This approach simply
re-writes the unconstrained optimisation as a constrained one, with a carefully
constructed sequence of auxiliary variables and envelopes, to deal with the
associated augmented Lagrangian; see 
\cite{parikh_proximal_2014,polson_proximal_2015, green_bayesian_2015} for
recent surveys.  Proximal algorithms have
achieved great success and provide a methodology for incorporating irregular
non-differentiable penalties; see \citep{masci_fast_2013} for applications in
the areas of computer vision and signal processing.

From a statistical perspective, DL models can be viewed as generalised linear
models (GLM \citet{davison_statistical_2003, dellaportas_bayesian_1993}) with
recursively defined link functions.  Traditional statistical models commonly
use shallow networks containing at most two layers or hierarchies. For example,
reduced rank regression can be viewed as a deep learning model with only two
layers and linear links. Support vector machines for classification
\cite{polson_data_2013} use a predictive rule based on a rectified linear (or
hinge) link.  Recent empirical evidence, however, suggests improved statistical
predictive performance with deep architectures of at least three layers.  Our
focus here will be on developing fast learning methods for deep multi-class
logistic models although our methods apply more generally to recurrent and
convolutional neural nets.  Although well known in the Neural Network and
Statistics literature \citep{knowles_nonconjugate_2011}, efficient estimation
of the cross-entropy/multinomial loss with regularization--outside of the
$\ell^2$-ridge penalty--has not been a mainstream area of research. See
\cite{madigan_bayesian_2005} and \cite{genkin_largescale_2007} for applications
to large scale multinomial logistic models. In general, the $\ell^2$-ridge penalty
is commonplace \citep{poggio_networks_1990, orr_regularization_1995}, mostly due to
its differentiability.   Our methods can therefore be seen as related to sparse
Bayes MAP techniques; see 
\cite{titterington_bayesian_2004, windle_efficient_2013,polson_proximal_2015}
for high dimensional data reduction.

Mainstream estimation within Deep Learning broadly revolves around gradient
descent methods.  The main variation in techniques arises from general
considerations for computational complexity and introduce more tuning
parameters \citep{ngiam_optimization_2011}.  Such considerations are the basis
for Stochastic Gradient Descent (SGD) and back-propagation.  For instance,
back-propagation uses the chain rule for the derivative of the composite of
activation functions. This can reduce the order-of-operations dramatically from
naive direct evaluation while maintaining high numerical accuracy; however,
this says little about the general difficultly in estimation of a non-linear
objective function.  Direct gradient methods can be poorly scaled for the
estimation of deep layer weights, in contrast to our proximal splitting
approach which overcomes this by providing a simultaneous block update of
parameters at all layers.  
The largest networks (e.g. \cite{dean_large_2012}) are currently trained using
asynchronous SGD.  \cite{farabet_learning_2013} discusses hardware approaches
to faster algorithms. Providing training methodologies is a very active field
of research.

The splitting techniques common in the proximal framework do exist in the AI
literature but we believe that their broad applicability and functional
coverage has mostly been overlooked and under-exploited. This is possibly due
to the aforementioned concerns of computational complexity that makes
stochastic gradient descent (SGD) and back-propagation methods so popular,
although it's quite possible that the number of iterations to step-complexity
can favor (parallel) proximal methods in some cases. We show that our augmented
ADMM approach is embarrassingly parallel with block updates for parameters and
auxiliary variables being directly available due to our carefully chosen
splitting procedure.

Traditional approaches to deep learning use Back-propagation
\citet{lecun_efficient_2012}, \cite{hinton_fast_2006},
\cite{hinton_reducing_2006} which rely on the chain rule for the derivatives of
the composite of the $L$ layers in the network.  We propose a proximal
splitting approach which also lends itself to the inclusion of a
non-differentiable regularisation penalty term so as to induce sparsity.
\cite{combettes_proximal_2011} detail multiple splitting methods in the context
of proximal operators as well as parallel proximal algorithms that handle many
splitting variables.  \cite{wang_nonlinear_2012} perform splitting in a similar
context as ours, but with an $\ell^2$ loss and quadratic barrier approach to
handle the equality constraints.  In contrast, we generalize to non-$\ell^2$ and
detail a general envelope approach that includes the well known augmented
Lagrangian.  Our general envelope approach allows one to utilize efficient
bounds, such as those found in \cite{bouchard_efficient_2007} and
\cite{knowles_nonconjugate_2011}. 

The rest of the paper is outlined as follows. Section~\ref{sec:deeplearning}
introduces deep learning predictors. We show that standard statistical models such
as principal components analysis, reduced rank and sliced inverse regression
can be viewed as shallow architecture models. We also motivate the addition
of further layers to aid in the regularised prediction problem. We
illustrate how proximal splitting methods solve a simple shallow architecture
model.  Section~\ref{sec:learning} describes the general deep learning problem
and its solution via proximal splitting methods. Supervised learning uses a
training dataset to estimate the parameters of each layer of the network.
We define the degrees of freedom of a deep learning
predictive rule which quantifies the trade-off between model fit and predictive
performance.  Section~\ref{sec:applications} provides a comparison of DL and
Neural Network (NN) models in a multi-class deep logistic classification model
for Fisher's Iris data. We also illustrate the convergence of our algorithm.
Finally, Section~\ref{sec:discussion} concludes with directions for future
research.

\section{Deep Learning Predictors}
\label{sec:deeplearning}

Let $y \in \mathcal{S} $ where $ \mathcal{S} = \Re^N $ for regression and 
$\mathcal{S} = \{1, \ldots, K\}$ for classification. Here $y$ denotes an
observed output associated with a high dimensional input/covariate variable
given by $\cX = \{X_i\}_{i=1}^N$ and $x_i \in \Re^{M} $.  The generic problem is
to find a non-linear predictor of the output $ \hat{y}(\cX,\cW)$ where $ \cW$ are
parameters which will be estimated from a training dataset of $N$ outputs and
inputs, denoted by $ \{ y_i,X_i \}_{i=1}^N $. 

When the dimensionality of $\cX$ is high, a common approach is to use a data
reduction  technique. This is achieved by introducing a low-dimensional auxiliary
variable $ z \in \Re^d $ where $ d \ll M $ and constructing a prediction rule specified by a
composition of functions, 
\begin{align*}
  \hat{y} &= f_1(f_2(\cX)) 
  \\
  &= f_1(z)
  \; \text{ where $z=f_2(\cX)$. }
\end{align*}
Now the problem of high dimensional data reduction is to find the $z$-variables
using training data and to estimate the layer functions $(f_1, f_2)$.  
One reason for the success of deep learning is the \textit{regularisation}
achieved through the hidden layer low-dimensional $z$ variable. A hallmark of
deep learning models is also the use of nonlinear layers.  As such,
one can view them as hierarchical nonlinear factor models or more
specifically as generalised linear models (GLM) with recursively defined
nonlinear link functions.

The key to a layering approach is to uncover the low-dimensional $z$-structure
in a way that doesn't disregard information about predicting the output $y$.
We will show that our splitting scheme naturally uses the hidden layer $z$-variables.

For example, \textit{Principal component analysis} (PCA) reduces $\cX$ using a
SVD decomposition \cite{wold_causal_1956}. This type of dimension reduction is
independent of $y$ and can easily discard $\cX$ information that is valuable for
predicting the output.  \textit{Sliced inverse regression} (SIR)
\cite{cook_dimension_1999}, \cite{cook_fisher_2007} overcomes this by
estimating the layer function $f_2$, independently of $f_1$, using data on both
$(y,\cX)$.  Deep learning takes this one step further and \textit{jointly}
estimates $f_1$ and $f_2$ in tandem using training data on both pairs
$(y,\cX)$.

A common approach is to introduce parameters, $W$, at each layer by convolution
with linear maps which we denote by $ f_1 ( W_1 z ) $ and $ f_2 (
W_2 x ) $ assuming centered variables.
Statistical models are traditionally shallow architectures with at most two layers.  
\textit{Reduced rank regression} corresponds to linear link functions 
with $ f_1 (z) = W_1 z $ where $ z = f_2(x) = W_2 x  $. The dimensionality
reduction then becomes an eigen-problem for $ W_1 $ and $W_2$.
\textit{Radial basis functions/kernel sliced inverse regression}
uses $f_1(z) = W \Phi(z)$ where $\Phi$ is a set of kernels/basis functions.
In many cases, we will also add
a penalty, $ \phi(\cW) $, for parameter estimation.  The
term $\phi(\cW)$ is a regularization term that imposes structure or effects a
favorable bias-variance trade-off in prediction. We need to be able to account
for non-smooth penalties such as  lasso $\phi(\cW) = \gamma \sum_{l=1}^L |W_l|$
or bridge penalty to induce sparsity, where $\gamma>0$ is a scaling
parameter that traces out a full regularisation path.

A deep learning predictor takes the form of an $L$-layered convolution rule of link functions where 
$$
\hat{y} = f_1(\dots f_{L-1}(z_L)\dots) \; \; \text{with} \; \; z_L = f_L(\cX) \; .
$$
We define the set of layer $z$-variables are given by $ z_l = W_l f_k ( z_{l+1}) $ 
assuming that the variables are centered. 

The original motivation for recursive deep learning predictors resides in the
seminal work of \cite{kolmogorov_representation_1957},
\cite{lorenz_deterministic_1963} and the extensions derived by
\cite{barron_universal_1993}, \cite{poggio_networks_1990} and
\cite{bach_breaking_2014-1}. See \cite{paige_bayesian_2001} for a Bayesian
approach to neural networks.  The motivation behind layered convolution maps
lies in the following completeness result. Given a Gaussian nonparametric
regression with any continuous function, $F(x)$  on $[0,1]^M$ and  $ \epsilon \sim N(0,I)$ \cite{poggio_networks_1990},  there exists
one dimensional link functions $f_l$ and $f_{l,m}$ such that
$$
y = F(\cX) + \sigma \epsilon 
\; \; \text{where} \; \; 
F(\cX) = \sum_{l=1}^{2M+1} f_l\left( \sum_{m=1}^M f_{l,m}(x_m) \right)
$$
This is a four layer network with a large number of units at the first layer.
On the practical side, one may wish to build a deep network with more layers
but less units at each stage. Section~\ref{sec:design} provides a model selection
approach to architecture design.

Neural network models can simply be viewed as \textit{projection pursuit regression} 
$F(x) = \sum_{m=l}^L f_l ( W_l x)$ with
the only difference being that in a neural network the nonlinear link functions,
$f_l$, are parameter dependent and learned from training data.
For example, a two-layer auto-encoder model with a sigmoid link function
uses a prediction rule given by functions of
the form $ \sum_l W_{1,l} \cdot \sigma \left( \sum_m W_{2,l,m} x_m + b_{m,0} \right)$. 
The $f_l$'s are learned via the sigmoid function of the linear combination.

The modeling intuition a deep learning architecture
can be found within a two layer system.
Suppose that we have a $\ell^2$ loss problem with a two layer rule
$$
\min_{W} \norm*{y - \hat{y}(\cX, \cW)}^2
\; \text{ where $\hat{y} = W_1 f_1(W_2 f_2(\cX))$} \; .
$$
This objective can be highly nonlinear. Finding the optimal parameters is challenging as traditional convex optimisation methods
deal only with sums rather than composites of objective functions. Our approach will be based on augmented
ADMM methods which still apply by splitting on the variables 
$z_1 = W_1 f_1 (z_2)$ and $z_2 = W_2 f_2 (\cX)$.
\cite{bertsekas_multiplier_1976} provides a general discussion of ADMM methods.

Now we need to solve an augmented Lagrangian problem of the form 
$$
\max_\cK \min_{\cW,\cZ} \cL(\cW,\cZ,\cK)
$$ 
with $\cK = \{\kappa_1, \kappa_2\}$
and $\cZ = \{z_1, z_2\}$ where
\begin{align*}
\cL(\cW,\cZ,\cK) = \sum_{i=1}^{N} \biggl\{
  &
  \norm*{y_i - z_{i,1}}^2 
  + \kappa_{i,1} \left( z_{i,1} - W_1 f_1(z_{i,2}) \right) 
  + \frac{\mu_{i,1}}{2} \norm*{ z_{i,1} - W_1 f_1(z_{i,2})}^2 
  \\
  &
  + \kappa_{i,2} \left( z_{i,2} - W_2 f_2(x_i) \right) 
  + \frac{\mu_{i,2}}{2} \norm*{z_{i,2} - W_2 f_2(x_i)}^2 
  \biggr\}
\end{align*}
for augmentation parameters $ ( \mu_{i1} , \mu_{i2} ) $.

Re-expressing in scaled Lagrangian form with 
$u_{i,j} = \kappa_{i,j}/\mu_{i,j}$, $j\in \{1,2\}$, gives
\begin{align*}
  \cL(\cZ,\cW,\cK) = \sum_{i=1}^{N} \biggl\{
  &
  \norm*{y_i - z_{i,1}}^2 
  \\ 
  &
  + \frac{\mu_{i,1}}{2} \norm*{z_{i,1} + u_{i,1} - W_1 f_1(z_{i,2})}^2 
  - \frac{\mu_{i,1}}{2} \norm*{u_{i,1}}^2 
  \\
  &
  + \frac{\mu_{i,2}}{2} \norm*{ z_{i,2} + u_{i,2} - W_2 f_2(x_i)}^2 
  - \frac{\mu_{i,2}}{2} \norm*{u_{i,2}}^2 
  \biggr\} \; .
\end{align*}
If we add an $\ell^2$-norm penalty for $\cW$, we obtain ridge regressions steps
in block $\cW$ updates.
The scaled Lagrangian saddle-point is solved via the iterative ADMM scheme
\cite{polson_proximal_2015}.
This solves the optimisation of a recursively defined set of link functions rather than
the traditional sum of convex functions.
See \cite{lewis_proximal_2008} for convergence analysis of these layered
optimisation problems.

An important feature of our ADMM update for the parameters $ \cW =(W_1, W_2)$
is that it happens in tandem. Moreover, the regularisation steps are
properly scaled by the current values of the functions of the augmented
$z$-variables.  On the other hand, a direct gradient-based approach such as
back-propagation uses first order information based on the composite derivative
$ ( f_1 \circ f_2 )^\prime $. This can easily lead to steps for the second
layer parameters, namely $W_2$, that are poorly scaled.  Back-propagation can 
be slow to converge as it makes small zig-zag steps in the highly multi-modal objective.
More efficient second order Hessian methods would require more information and computation.
\citet{martens_learning_2011} develops a Hessian-free Lagrangian
approach that is based on an envelope that uses the derivative of the
composite map $ l \circ f $ where $l$ denotes the model measure of fit and $f$
a layer function.

\begin{table}
  \begin{center}
  \caption{\label{tab:links} Link (activation) functions. Typical $L_p$-norms
    are $p=1$ (lasso), $p=2$ (ridge) or $\infty$ (max-norm).
    RectLU/hinge norm is related to lasso via $ \max (u,0) = \frac{1}{2} ( |u| + u ) $, see 
    \cite{polson_data_2011}. Max-pooling is the sum
    of a hinge and lasso norms,
    $ \max( |u_1| , |u_2| ) = \max( |u_1| - | u_2| , 0 ) + |u_2| $. }
  \vspace{1pc}
  \begin{tabular}{r r}
    & $f_k(u)$  \\
    \hline
    linear & $A u + b$  \\
    sigmoid & $(1 + e^{u} )^{-1}$  \\
    softmax & $ e^u / \sum_{k=1}^K e^{u_k}$ \\
    tanh & $2 (1 + e^{u} )^{-1}-1$ \\
    log-sum-exp & $ \log \sum_i e^{u_i} $ \\
    $\ell^p$-norm & $\left( \sum_i |u_i|^p \right)^{\frac{1}{p}}$ \\
    rectLU/hinge &  $\max(0,u)$  \\
    max-pooling & $\max\{|u_{I_i}| , |u_{I_j}|\}$ 
  \end{tabular}
  \end{center}
\end{table}

Whilst our augmented Lagrangian uses an $\ell^2$-barrier, we can use
general metrics and still achieve convergence \cite{bertsekas_multiplier_1976}.
A computationally attractive approach would be to match the type of ADMM barrier
with the nonlinearity in the link function. We leave this for future research.

We now turn to our Deep Bayes learning framework. Bayes provides a way of
unifying the stochastic and algorithmic approaches to data
\cite{breiman_statistical_2001}.

\section{Deep Bayes Learning}
\label{sec:learning}

A deep learning predictor, 
$\hat{y}(x,\cW) \in \Re^N$, depends recursively on a set of link functions, 
$ f_k , 1 \leq k \leq L$,  defined by
$$
\hat{y}_i \defeq \hat{y}(x_i, \cW) = W_1 f_1 \left( W_2 f_2(
  \ldots(f_{L-1}(W_L X_i  + b_L))\dots) + b_2 \right) \; .
$$
The link or activation functions $f_l: \Re^{N_{l}} \to \Re^{N_l}$ are
pre-specified and part of the architecture design.  Specific choices includes
\textit{sigmoidal}; \textit{softmax};
\textit{tanh}; \textit{rectLU} or \textit{log-sum-exp} see Table 1.  
We use $L$ to denote the number of layers in the network.  The parameters 
$b_l \in \Re^{N_l}$, with $1 \leq l \leq L$ and 
$b_1 = \0$, are off-sets or bias parameters. The parameters $ \cW$ can be decomposed as  
$$
\mathcal{W}=(W_1, \ldots, W_L)
\text{ for } W_l \in \Re^{N_l \times N_{l+1}} \text{ for $1 \leq l < L$ and } N_L = M,
$$ 
To construct a posterior distribution, $p( \cW|y)$, we use
a prior distribution,
$p(\cW) \propto \exp \left(-\phi(\cW) \right)$, where $ \phi(\cW) $ is
a regularisation penalty.

Bayes rule yields the posterior distribution over parameters given data, namely 
\begin{align*}
p(\cW|\cX,y) & \propto p(y | \hat{y}(\cX,\cW)) p(\cW)\\ 
  & \propto \exp \left(-\log p(y|\hat{y}(\cX, \cW) ) - \phi(\cW) \right) \; .
\end{align*}

The deep learning estimator, $ \hat{W}$,  is a Bayes MAP estimator and leads to
a prediction rule
$$
\hat{y} ( \cX , \hat{W} ) \; \; \text{where} \; \; \hat{W} \defeq 
\argmax_{\cW} p ( \cW | \cX , y ) \; . $$
Maximising the posterior distribution corresponds to finding the
$\argmin$ of the deep learning objective function.  Taking a Bayesian perspective
allows the researcher to characterise the full posterior
distribution, $p(\cW|y)$, typically, via simulation methods such as MCMC. This
can aid in finding parameter uncertainty estimates and in determining efficient prediction rules.

The log-likelihood can also be interpreted as gauging
the accuracy of a prediction, $\hat{y}$, of outcome, $y$. 
The underlying probability model, denoted by $p(y| \hat{y})$, determined this fit via 
$$
l(y, \hat{y} ) = - \log p( y | \hat{y} ) \;.
$$
The function $l(y, \hat{y})$ is typically a smooth
function such as an $\ell^2$-norm $ \norm{y - \hat{y}}^2 $ or a cross-entropy
metric $ l(y,\hat{y}) = y \log \hat{y} $ when dealing with classification. The
statistical interpretation is as a negative log-likelihood, in machine
learning a cost functional. The major difficulty in implementing deep learning models is the high degree of
nonlinearity induced by the predictor, $ \hat{y} ( \cW , \cX ) $, in the likelihood.

We will pay particular attention to the multi-class deep logistic learning model. The
first layer is given by $\sigma(W x) = e^{W x}/\sum_k^K e^{W_{k} x}$
which defines the vector sigmoid function $\sigma(x)$. 

\begin{Exa}[Deep Logistic Learning]

Suppose that our
observations $y_t$ represent a multi-class $1$-of-$K$ indicator vector,
which we equate with class $k$ via $y=k$ for $1 \leq k \leq K$. Given a set of
deep layer link functions with $z_l=f_l(\cdot)$, we have a predictive rule 
$$
\hat{y}(z, \cW) = p(y=k|z,\cW) = \sigma_k(W_{1} z) \; .  
$$
The negative log likelihood is given by
\begin{align*}
  l(y_i, \hat{y}_i) = \log p(y_i|\hat{y}_i) 
  &= -\log \prod_{k=1}^{K} (\hat{y}_{i,k})^{y_{i,k}}
  \\ 
  &= -\sum_{k=1}^{K} y_{i,k} \log \hat{y}_{i,k}  \; .
\end{align*}
Minimising the cross-entropy cost functional is therefore equivalent to a
multi-class logistic likelihood function. \cite{genkin_largescale_2007} and 
\cite{madigan_bayesian_2005} provide analysis of large-scale
multinomial logit models with shallow architectures.
\end{Exa}

The basic deep learning problem is supervised learning with a training dataset
$(y_i , X_i)_{i=1}^N$. We find the deep network structure via an
optimisation of the following form
\begin{equation}
\begin{gathered}
  \argmin_{\cW} 
  \left\{ l(y_i , \hat{y}(x_i, \cW)) + \phi(\cW) + \phi(\cZ) \right\} 
  \\
  \text{where } 
  \hat{y}(x_i, \cW) = W_1 f_1 \left( W_2  
  f_2(\ldots( f_{L-1}(W_L X_i + b_L))\dots) + b_2 \right)
  \label{eq:primal_obj}
\end{gathered}
\end{equation}
Again $l(y,\hat{y})$ is a measure of fit depending implicitly on some observed data
$y$ and a prediction rule $ \hat{y}(\cW,\cX) $.  Here $y$ denotes an
$N$-vector of outcomes and $\cX$ a corresponding set of $N$-many $M$-vector
characteristics; for example, pixels in an image or token counts 
in a topic model for document classification.

To solve the deep learning objective function with non-differentiable
regularisation penalties we use an auxiliary latent variable scheme
in the context of an augmented Lagrangian.
This allows us to write the deep learning optimisation as a constrained problem
\begin{equation}
  \begin{aligned}
    \argmin_{\cZ, \cW} & \; \sum_{i=1}^N l(y_i , z_1) + \phi(\cW) + \phi(\cZ) 
  \\
  \text{where }& z_{l} = W_l f_{l}(z_{l+1}) + b_l \quad 1 \leq l < L
  \\
  & z_{L} = W_L x_i + b_L
\end{aligned}
\label{eq:split_objective}
\end{equation}
Through variable splitting we can introduce latent auxiliary variables, 
$z_{l} \in \Re^{N_l}$, where $f_l(z_{l+1}) : \Re^{N_{l+1}} \to \Re^{N_{l+1}}$.
Notice that we also allow for regularisation of the $ \cZ$ variables so as to
include sparsity.

Our approach follows the alternating pattern of ADMM and Douglas-Rachford
type algorithms, which for a split objective given by
$$
\min f(w) + g(z) \text{ where } A w - B z = 0 
$$
performs the following iterations
\begin{align*}
  w^{(t+1)} &= \argmin_w \left\{ 
    f(w) + \frac\mu2 \norm*{w - (z^{(t)} - u^{(t)})}^2
  \right\}
  \\
  z^{(t+1)} &= \argmin_z \left\{ 
    g(z) + \frac\mu2 \norm*{z - (w^{(t+1)} + u^{(t)})}^2
  \right\}
  \\
  u^{(t+1)} &= u^{(t)} + A w^{(t+1)} - B z^{(t+1)}
  \;.
\end{align*}
for some $\mu > 0$.

Roughly speaking, variable splitting enables one to formulate an
iterative solution to the original problem that consists of simple first-order
steps, or independent second-order steps, since it ``decouples'' the
functions, which in our case are the penalty, layers, and loss
functions.  Now, potentially simpler minimization steps on the splitting
variables and observation are combined, often in a simple
linear way, to maintain the ``coupling'' that exists in the original problem.
In such a setting, one can craft sub-problems--through the choices of
possible splittings--that have good conditioning or that suit the computational
platform and restrictions, all without necessarily excluding the use of
standard and advanced optimization techniques that may have applied to the
problem in its original form.  

For instance, the same Newton-steps and back-propagation techniques can be
applied on a per-layer basis.  For that matter, one can choose to split at the
lowest layer (i.e. between the loss function and layers above) and allow $f_1$
to be a complicated composite function.  This separates the
problem of minimizing a potentially complicated loss from a composition of
non-linear functions.  In general, steps will still need to be taken in the
composite $f_1$, but again, currently established methods can be applied.

Since splitting can occur at any level, a natural question is  
at which layer should the splitting occur.  If there was  
a penalty function across splitting variables, then standard regularization
paths could be computed in fairly low dimension (minimum 2, for a weight on $\cW$
and $\cZ$).  This provides a potentially smooth scaling across the
layer dimension and/or number of layers (see the regressor selection problems 
in \cite[Section 9.1.1]{boyd_distributed_2011}).  By simply re-applying the
strict equality constraints between splitting variables in some layers, 
e.g. through an indicator penalty $\phi(\cZ)$, one could effectively emulate certain
model designs; thus, through a structured approach to regularizing
these terms one can perform a type of model selection.  This approach may
also explain--and emulate--the effect of drop-out in a non-stochastic way. 

The recursive nature of defining layers naturally leads to the construction of
$z$-variables and in Section~\ref{sec:splitting} and~\ref{sec:prox_admm}
we show how this provides an under-exploited relationship between these methods
and deep learning.

\section{Proximal Splitting}
\label{sec:splitting}

Let's start by considering the augmented Lagrangian, $\cL(\cZ,\cW, \cK)$, 
which takes the following general form for observations $\{(y_i, X_i)\}_{1:N}$
\begin{equation}
\begin{split}
  \max_\cK \min_{\cZ, \cW} \sum_{i=1}^N \biggl\{ 
  & \phi(\cW)  + l(y_i, z_{i,1}) 
  \\ 
  &+ \kappa_{i,1}^\top \left( z_{i,1} - W_1 f_1(z_{i,2}) - b_{1}\right) 
  + \frac{\mu_{i,1}}{2} \norm{z_{i,1} - W_1 f_1(z_{i,2}) - b_{1}}^2
  \\ 
  & + \sum_{l=2}^{L-1} \left( 
    \kappa_{i,l}^\top \left( z_{i,l} - W_l f_l(z_{i,l+1}) - b_{l}\right)
    + \frac{\mu_{i,l}}{2} \norm{ z_{i,l} - W_l f_l(z_{i,l+1}) - b_{l}}^2
    \right)
  \\ 
  & + \kappa_{i,L}^\top \left( z_{i,L} - W_L x_i - b_L \right) 
    + \frac{\mu_{i,L}}{2} \norm{ z_{i,L} - W_L x_i - b_L }^2 
    \biggr\}
\end{split}
\label{eq:aug_lagrange}
\end{equation}  
where $\kappa_{i,l} \in \Re^{N_l}$ and $\kappa_{i,l} \in \cK$ 
are Lagrange multipliers and $\mu_{i,l} \in \Re_+$.
The form here does not easily highlight the role of each
term across observations and layers, nor the relationship between terms. As a result, in what follows,
we recast the Lagrangian into
forms that clarify the relationship. 


We make extensive use of vectorization, so many of the result are obtained by
using the following identities
\begin{align*}
  \vec(AB) &= (I \otimes A) \vec(B)
   = (B^\top \otimes I) \vec(A)
\end{align*}
which, in our case, give 
\begin{align*}
  \vec(W_l f_l(Z_{l+1})) &= (I_N \otimes W_l) f_l(z_{l+1})
   = (f_l(Z_{l+1})^\top \otimes I_{N_l}) w_l
\end{align*}
where $ w_l = \vec(W_l) $, for the stacked by observation terms
\begin{gather*}
  Z_l = (z_{1,l} \dots z_{N,l}) \in \Re^{N_l \times N} 
  \text{ and }
  z_l = \vec(Z_l) \in \Re^{N_l \cdot N}  
  \\ 
  X = \left(x_1 \dots x_N \right) \in \Re^{M \times N}
  \text{ and }
  \mu_l = \left(\mu_{1,l} \dots \mu_{N,l} \right)^\top \in \Re^N_{+} 
\; .
\end{gather*}
We also extend $W_{l}$ and $f_l(z_{i,l+1})$ with 
$\tN_l = N_l + 1$ and
\begin{gather*}
\tW_{l} = 
\begin{pmatrix}
  b_l & W_l
\end{pmatrix}
\in \Re^{N_l \times \tN_{l+1}}
,\quad
\tf_l(z_{i,l+1}) =
\begin{pmatrix}
  1
  \\
  f_l(z_{i,l+1})
\end{pmatrix}
\in \Re^{\tN_{l+1} \times 1}
\\
\tw_{l} = \vec(\tW_l) 
\in \Re^{N_l \cdot \tN_{l+1}}
,\quad
\tf_l(Z_{l+1}) =
\begin{pmatrix}
  \1^\top_{N} 
  \\
  f_l(Z_{l+1})
\end{pmatrix}
\in \Re^{\tN_{l+1}\times N}
\end{gather*}
which includes $f_L(z_{i,L+1}) \defeq x_i$.
This means that the $b_l$ terms are members of the primal variable set $\cW$.

We introduce the scaled Lagrange multipliers 
$$
u_l^\top = \left(\frac{\kappa_{1,l}^\top}{\mu_{1,l}} 
  \dots \frac{\kappa_{N,l}^\top}{\mu_{N,l}}\right) \in \Re^{N_l \cdot N}
\;.
$$ 
We then obtain
\begin{gather}
  \begin{aligned}
  \max_\cU \min_{\cW, \cZ} \biggl\{ 
  & \phi(\cW) + L(y, z_{1}) + \sum_{l=1}^{L} \frac{1}{2} 
    \norm{ z_{l}  
      - \left(I_N \otimes W_l\right)  f_l(z_{l+1}) 
      - \left(\1_N \otimes b_{l}\right)
      + u_{l}
    }^2_{\Lambda_{\mu_l}}
  \\ 
  & - \frac12 \sum_{l=1}^{L} \left( \norm{u_1}^2_{\Lambda_{\mu_1}} \right)
  \biggr\}
  \end{aligned}
  \nonumber
  \\
  =
  \begin{aligned}
  \max_\cU \min_{\cW, \cZ} \biggl\{ 
   & \phi(\cW) + L(y, z_{1}) + \sum_{l=1}^{L} \frac{1}{2} 
    \norm{ z_{l}  
      - \left(I_N \otimes \tW_l\right)  \tf_l(z_{l+1}) 
      + u_{l}
    }^2_{\Lambda_{\mu_l}}
  \\
  & - \frac12 \sum_{l=1}^{L} \left( \norm{u_1}^2_{\Lambda_{\mu_1}} \right) 
    \biggr\}
  \label{eq:reduced_lagrange}
  \end{aligned}
\end{gather}
where $ I_N $ is the $ N \times N $ identity matrix and
$$
\Lambda_{\mu_l} = \Diag(\mu_l \otimes \1_{N_l}) = \bigoplus_{i=1}^N \mu_{i,l} I_{N_l}
  \in \Re_+^{(N \cdot N_l) \times (N \cdot N_l)} \;,
$$ 
inner-product norm 
$\norm{x}^2_{\Lambda} \defeq x^\top \Lambda x$ and with $\otimes$ denoting the
Kronecker product and $\bigoplus$ the direct sum.  
$\Lambda_{\mu_l}$ is a block diagonal, so it can be factored in
``square root'' fashion.

Naturally, operations across layers $l$ can also be vectorized.  
First, let $N_{w} = N \sum_{n=1}^L \tN_n$, $N_z = N \sum_{n=1}^L N_n$ and
\begin{gather*}
  z^\top = \left(z_1^\top, \dots, z_L^\top\right) 
    \in \Re^{N_z}
  \\
  u^\top = \left(u_1^\top, \dots, u_L^\top\right) 
    \in \Re^{N_z}
  \\
  \Lambda_{\mu} = \bigoplus_{n=1}^L \Lambda_{\mu_n}
    \in \Re^{N_z \times N_z} \; .
\end{gather*}
The linear first-order difference maps are defined by
\begin{equation}
\begin{aligned}
  \Delta_{\tw} &: \Re^{N_z} \to \Re^{N_z} 
  \\ 
  \Delta_{\tw} &\defeq I_{N_z} -
  \begin{pmatrix}
    0 & \left(I_N \otimes \tW_1 \right) & 0 & \dots & 0
    \\ 
    \vdots & 0 & \left(I_N \otimes \tW_2 \right) & 0 & \vdots
    \\
    \vdots & \vdots & 0 & \ddots & 0
    \\
    0 & 0 & \dots & 0 & \left(I_N \otimes \tW_L \right)  
  \end{pmatrix} \tf
  \\
  &= I - \Omega_{\tw} \tf
\end{aligned}
\label{eq:delta_w}
\end{equation}
where the $0$ matrices match in dimension, 
$\Omega_{\tw} \in \Re^{N_z \times (N_w + N \cdot M)}$ and 
$$
\tf \circ z \defeq \left(z^\top_1, \tf_1(z^\top_2), \dots, \tf_{L-1}(z^\top_L), 
  \vec(X)^\top \right)^\top 
\;.
$$
Now, with $P_1 z = z_1$, \eqref{eq:reduced_lagrange} becomes
\begin{equation}
  \max_u \min_{\tw, z} \left\{ 
  \phi(\tw) + L(y, P_1 z) 
  + \frac12 \norm{ \Delta_{\tw} z + u}^2_{\Lambda_{\mu}}
  - \frac12 \norm{u}^2_{\Lambda_{\mu}} 
  \right\}
  \label{eq:full_reduced_lagrange_z}
\end{equation}
In terms of $\tw$, our problem is
\begin{equation}
  \max_u \min_{\tw, z} \left\{ 
   \phi(\tw) + L(y, P_1 z) 
  + \frac{1}{2} 
    \norm{ \Delta_z \tw - z - u}^2_{\Lambda_{\mu}}
  - \frac12 \norm{u}^2_{\Lambda_{\mu}} 
  \right\}
  \label{eq:full_reduced_lagrange_w}
\end{equation}
where
\begin{equation}
  \Delta_z \defeq \bigoplus_{n=1}^L 
    \left({\tf}_n(Z^\top_{n+1}) \otimes I_{N_n} \right) 
    \in \Re^{N_z \times N_w} \; .
  \label{eq:delta_z}
\end{equation}
Note the relationship between the two operators, i.e.
$$
\left(I+\Delta_{\tw}\right) z = \Delta_z \tw 
\;.
$$

Equations~\eqref{eq:full_reduced_lagrange_z}
and~\eqref{eq:full_reduced_lagrange_w} provides a simple form that shows
how our problem differs from the problems commonly considered in the basic
operator splitting literature, in which ADMM, Douglas-Rachford and the
inexact-Uzawa techniques are developed.  In these cases the general constraint is
usually given as $A w + B z = c$, for linear operators $A$ and $B$.  Our
problem involves an operator, $\Delta_{\tw}$, that introduces a multiplicative
relationship between the primal variable $\tw$ and the dual $z$.  
This form could be interpreted as--or related to--a bi-convex problem 
(for convex $f_l$, naturally), especially when $\Delta_{\tw}$ is bi-affine
\citet{boyd_distributed_2011}.

\subsection{Proximal Operators and ADMM}
\label{sec:prox_admm}

The proximal operator is defined by
\begin{align*}
  \prox_{\Lambda g}(x) 
    \defeq \argmin_z \left\{ g(z) + \frac{1}{2} \norm{z-x}^2_\Lambda \right\}    
\end{align*}
for positive definite $\Lambda$.  
Normally, $\Lambda = \lambda I$ and the operator reduces to
\begin{align*}
  \prox_{\lambda g}(x) 
    \defeq \argmin_z \left\{ g(z) + \frac{\lambda}{2} \norm{z-x}^2 \right\}    
    \;.
\end{align*}
When $\Lambda$ is diagonal and positive, such as $\Gamma_{\mu_l}$ above,
one could simply use its inverse to rescale the terms in the problem (effectively
$g(z) \to g(\Lambda^{-1/2} z) \defeq \tilde{g}(z)$ and 
$x \to \Lambda^{-1/2}x$); however, we use the matrix inner-product norm for
notational convenience and the implication of wider applicability.

The proximal operator enjoys many convenient properties,
especially for lower semi-continuous, convex $g$, and has strong connections
with numerous optimization techniques and fixed-point methods 
\citep{boyd_convex_2009,combettes_proximal_2011}.  

The form of \eqref{eq:reduced_lagrange} in $z_l$ reflects the definition
of the proximal operator, and, after some manipulation, the same is true
for $\tW_l$.  This
allows us to apply the iterative techniques surrounding proximal operators
in what follows.

One advantage of our approach is its embarrassingly
parallel nature. The augmented Lagrangian leads to a block update of 
$(\cW | \cZ, \cU)$. For example, if we add the traditional $\ell^2$-ridge penalty
we have a Bayes ridge regression update for the block $(\tW_1 , \ldots , \tW_L)$.
Proximal algorithms can also be used for non-smooth penalties.  Our method is
also directly scalable and the vectorization of our method allows
implementation in large scale tensor libraries such as {\tt Theano} or {\tt Torch}.

\begin{enumerate}
  \item
    \label{z_step}
    \noindent{$\cZ^{(t)}$ given $\left\{\cZ^{(t-1)}, \cW^{(t-1)}, \cU^{(t-1)}\right\}$}

    The problem, in vectorized form, is
    \begin{align*}
      \argmin_{z} \left\{ 
      L(y, P_1 z) + \frac12 \norm{ \Delta_{\tw} z + u}^2_{\Lambda_{\mu}}
      \right\}
    \end{align*}
    which, given the design of $\Delta_{\tw}$, is not a simple proximal
    operator.  Even if $\Delta_{\tw}$ resulted in a simple diagonal matrix, 
    the proximal operator of $L(y, P_1 z)$ may not be easy to evaluate. 
    Regardless, if this minimum is found using proximal approaches, 
    it would likely be through another phase of splitting, be it
    ADMM for the subproblem, or forward-backward/proximal gradient iterations.

    To illustrate the forward-backward approach we let 
    $F(z) = \frac12 \norm{ \Delta_{\tw} z + u}^2_{\Lambda_{\mu}}$ and note that
    the Jacobian matrix, $\D F(z)$, is given by
    \begin{equation}
    \begin{aligned}
      \D F(z) &=
      \left(I - \Omega_{\tw} \frac{\partial \tf}{\partial z^\top} \right)^\top
      \Lambda_\mu \left(\Delta_{\tw} z^{(t)} + u^{(t)} \right)
      \\
      &=
      \left(I - \Omega_{\tw} \frac{\partial \tf}{\partial z^\top} \right)^\top
      \Lambda_\mu \left(z^{(t)} - \Omega_{\tw} \tf(z^{(t)}) + u^{(t)} \right)
    \end{aligned}
    \label{eq:F_grad}
    \end{equation}
    and use gradient step (note that $\D f(z) = \nabla^\top f(z)$)
    \begin{align}
      s &= z^{(t)}- \gamma \nabla F(z^{(t)}) 
      \\
      z^{(t+1)} &= s + P^\top_1 \left( \prox_{\gamma L(y, \cdot)} 
          \left( P_1 s \right) - P_1 s \right) 
        \label{eq:z_L_fb}
    \end{align}
    Simple forward-backward can't be expected to work well for all moderate
    to high-dimensional problems, and especially not for functions with more extreme
    non-linearities.  Given the composite nature of $F(z)$, the sensitivity and 
    condition of this problem could vary drastically, so one will have to tailor
    their approach to account for this.  

  \item
    \label{w_step}
    \noindent{$\cW^{(t)}$ given $\left\{ \cZ^{(t)}, \cW^{(t-1)}, \cU^{(t-1)}\right\}$}

    From \eqref{eq:full_reduced_lagrange_w}, the problem is
    \begin{align*}
      \argmin_{\tw} \left\{ 
       \phi(\tw)  + \frac{1}{2} \norm{ \Delta_z \tw - (z + u)}^2_{\Lambda_{\mu}}
      \right\}
      \;.
    \end{align*}

    It is easier to see the relationships between terms when operating on
    a single layer, and since this sub-problem is separable by layer, we 
    proceed conditionally on $l$.  Let $\tf_{i,l} \defeq \tf_l(z_{i,l+1})$ then
    \begin{equation}
    \begin{aligned}
      \argmin_{\tw_l} & \biggl\{      
        \phi(\tw_l) +  
        \sum_{i=1}^{N} \frac{\mu_{i,l}}{2} \norm*{
          \left(\tf_{i,l}^\top \otimes I_{N_l}\right) \tw_l 
          - \left(z_{i,l}+u_{i,l}\right)}^{2}
      \biggr\}
      \\
      &= 
      \prox_{\Lambda_{w_l} \phi}\left(
        \Lambda_{w_l}^{\dagger} 
        \left(\tf_l(Z_{l+1}) \Diag(\mu_l) \otimes I_{N_l}\right)
        \left(z_l + u_l\right)
      \right)
      \\ 
      &= 
        \prox_{\Lambda_{w_l} \phi}\biggl\{
          \Lambda_{w_l}^{\dagger}
          \vec\bigl(\left(Z_l + U_L\right) \Diag(\mu_l) \tf_l(Z^\top_{l+1})\bigr) 
        \biggr\}
    \end{aligned}
    \label{eq:W_l_prob}
    \end{equation}
    where 
    \begin{align*}
      \Lambda_{w_l} &= \sum_{i=1}^{N} \mu_{i,l} \left(\tf_{i,l} \otimes I_{N_l}\right) 
      \left(\tf_{i,l}^\top \otimes I_{N_l}\right) =
        \sum_{i=1}^{N}\mu_{i,l} \left(\tf_{i,l} \tf_{i,l}^\top \otimes I_{N_l}\right)  
        \\
        &= \tf_l(Z_{l+1}) \Diag(\mu_l) \tf_l(Z_{l+1})^\top \otimes I_{N_l}
    \end{align*}
    and $\Lambda_{w_l}^{\dagger}$ is a right pseudo-inverse. 
    See Appendix~\ref{app:w_l_steps} for details.

    The resulting proximal problem involves a quadratic in $\tw_l$ that is no
    longer strictly diagonal in its squared term and, thus, isn't necessarily a
    simple proximal operator to evaluate.  The operator splitting that
    underlies ADMM, Douglas-Rachford and similar techniques is a common
    approach to this type of problem.  Also, if full decompositions (e.g. SVD,
    eigen) of $\Lambda_{w_l}^{\dagger}$ are reasonable to compute at each
    iteration, one could proceed by working in transformed $\tw_l$ coordinates;
    however, the transform will induce a dependency between components of
    $\tw_l$ in $\phi$, which may require proximal solutions that are themselves
    difficult to compute.  The composite methods in
    \cite{argyriou_learning_2013,chen_primaldual_2013} are designed for such
    transformed problems, at least when the transform is positive-definite, but
    given the dependency on $\tf_l$, such conditions are not easy to guarantee
    in generality.  

    Other approaches for solving this sub-problem are forward-backward
    iterations and ADMM; each approach should be considered in light of $\tf_{i,l}$.  
    A forward-backward approach may be trivial to implement,
    especially when $\tf_{i,l}$ has a known bound, but convergence can be
    prohibitively slow for poorly conditioned $\Lambda_{\tw}$.  
    Some ADMM approaches can lead to faster convergence rates, but at the cost
    of repeated matrix inversions, which--for structured matrix problems--may
    be trivial. 

    For example, a forward-backward approach for lower semi-continuous, convex
    $\phi$ would consist of the following fixed-point iterations
    $$
    \tw_l = \prox_{\lambda_{w} \phi}\left(\tw_l 
      - \lambda_{w} \nabla F(\tw_l)\right)
    $$
    where $\lambda_w \geq 0$ and
    $$
    F(\tw_l) = \frac{1}{2} 
    \tw_l^\top \Lambda_{w_l} \tw_l - \tw_l^\top d_{w_l} + c_{\tw_l} 
    $$
    with $\xi_{i,l}= z_{i,l}+u_{i,l}$ and 
    \begin{gather*}
      d_{\tw_l} = \sum_{i=1}^{N} \left(\mu_{i,l} \tf_{i,l} \otimes I_{N_l}\right) \xi_{i,l}
      ,\\
      c_{\tw_l} = \frac12 \sum_{i=1}^{N}\mu_{i,l}\norm*{\xi_{i,l}}^2 
    \end{gather*}
    If $\nabla F(\tw_l)$ is Lipschitz continuous with constant $\gamma_w$, then
    $\lambda_w \geq \gamma_w$; otherwise, line-search can be used to find
    a sufficient $\gamma_w$ at every iteration.

    An ADMM approach could result in a single primal proximal step in $F$, which
    involves inversions of a quantity like $\lambda I + \Lambda_{w_l}$, and, given
    the form of $\Lambda_{w_l}$, may be possible to compute via the well-known
    identity
    $$
    \left(I + u v^\top\right)^{-1} = I - \frac{u v^\top}{1+ v^\top u}  
    \;.
    $$

  \item
    \label{u_step}
    \noindent{$\cU^{(t)}$ given $\left\{\cZ^{(t)}, \cW^{(t)}, \cU^{(t-1)}\right\}$}
    This involves the standard cumulative error update for the augmented Lagrangian,
    which is 
    \begin{equation}
    \begin{aligned}
      u^{(t)} = u^{(t-1)} + \Delta^{(t)}_{\tw} z^{(t)}
    \end{aligned}
    \label{eq:u_update}
    \end{equation}
\end{enumerate}

Our approach requires a ``consensus'' across observations in Step~\ref{w_step},
but the remaining steps are free to be processed asynchronously in observation
dimension $N$.  The structure of both $\Delta_{\tw}$ and $\Delta_z$ essentially
determine the separability, since they each act as ``correlation'' matrices
between $z$ and $\tw$.
Observing \eqref{eq:delta_w}, we see that $\Omega_{\tw}$ sums ``horizontally'' across
layers, but in independent blocks of observations.   


\subsection{Model Selection and Architecture Design}
\label{sec:design}

One advantage of viewing a deep learning predictor as a Bayes MAP estimator is
that we can seamlessly apply Bayesian
model selection tools to determine optimal architecture design.  
We will provide a model selection criterion for choosing between different link
functions, size of the number of units together with
the number of layers. Given the probabilistic data generating model, 
$p(y | \hat{y}) $ and a deep learning estimator $\hat{y}(\cX, \hat{\cW_L})$
based on $L$ layers, we propose the use of an information criteria (IC) to
gauge model quality. Such measures like AIC, corrected AIC
\cite{hurvich_bias_1991}, BIC and others \cite{george_calibration_2000} have a
long history as model selection criteria. From a Bayesian perspective, we can
define the Bayes factor as the ratio of marginal likelihoods and also provide
an optimal model averaging approach to prediction. 

An information measure for a deep learning predictor, $\hat{y} $, or
equivalently a given architecture design, falls into a class of the form 
$$
IC(\hat{y}(L)) = - 2 \log p( y | \hat{y}(\cX, \hat{\cW_L} )  + c \cdot \text{df} \;.
$$
where $ \text{df} $ is a measure of complexity or so-called 
\textit{degrees of freedom} of a model and $c$ is its cost. The degrees of
freedom term can be defined as 
$ df \defeq \sigma^{-2} \sum_{i=1}^N \operatorname{Cov}(y_i , \hat{y}_i ) $
where $ \sigma^2$ is the model estimation error.

The intuition is simple--the first term assesses in-sample predictive fit.
However, over-fitting is the curse of any nonlinear high dimensional
prediction or modeling strategy and the second term penalises for the complexity
in architecture design--including the nonlinearities in the links, number of
units and layer depth. The combination of terms provides a simple metric for
the comparison of two architecture designs--the best model provides the highest
IC value.

For suitably stable predictors, $\hat{y}$, \cite{stein_estimation_1981} provides
an unbiased estimator of risk using the identity 
$\text{df} = \mathbb{E} \left(\sum_{i=1}^N \partial \hat{y}_i/\partial{y}_i \right ) $.
Given the scalability of our algorithm, the derivative 
$ \partial \hat{y} / \partial y $ is available using the chain
rule for the composition of the $L$ layers and computable using standard tensor
libraries such as {\tt Torch}.

\cite{efron_estimating_1983} provides an alternative motivation for model selection 
by directly addressing the trade-off between minimising out-of-sample
predictive error and in-sample fit.  Consider a nonparametric
regression under $\ell^2$-norm. The in-sample mean squared error is
$ \text{err} = ||y- \hat{y}||^2 $ and the out-of-sample predictive MSE is 
$  \text{Err} = \mathbb{E}_{ y^\star} \left ( ||y^\star - \hat{y}||^2 \right )
$ for a future observation $ y^\star $.
In expectation we then have
$$
\mathbb{E} \left ( \text{Err} \right ) = 
\mathbb{E} \left ( \text{err} + 2 \var ( \hat{y},y ) \right )
$$  
The latter term can be written in terms of $ \text{df} $ as a covariance.    
 Stein's unbiased risk estimate then becomes
$$
  \widehat{ \text{Err} } = ||y- \hat{y}||^2 
  + 2\sigma^2 \sum_{i=1}^{n} \frac{\partial \hat{y}_i}{\partial y_i}
  .
$$
Models with the best predictive MSE are favoured.
This approach also provides a predictive MSE criterion for optimal
hyper-parameter selection in the prior regularisation penalty $ \phi( \cW ) $
and allow the researcher to gauge the overall amount of regularisation
necessary to provide architectures that provide good predictive rules.
In contrast, the current state-of-the-art is to use heuristic rules such as dropout.  

\section{Applications}
\label{sec:applications}

To illustrate our methodology, we provide an illustration of multi-class deep
logistic learning. We use logit/softmax activation functions which allows us to
employ some efficient quadratic envelopes to linearize the functions in the
proximal sub-problems within our algorithm.  We use Fisher's Iris data to
illustrate the comparison between DL models and traditional classification
algorithms such as support vector machines.

\subsection{Multi-Class Deep Logistic Regression}
\label{sec:multinom}

Suppose that we have a multinomial
loss with $K$-classes, where the observations are $Y \in \Re^{K \times N}$,
and $f_l(z) = \sigma(z)$ are all logistic functions.
Now, \eqref{eq:z_L_fb} has an explicit form as
\begin{align*}
  L(Y, Z_1) &= 
  \sum_{i=1}^N \left\{\log\left(\sum_{k=1}^{N_1}e^{Z_{k,i,1} }\right) 
    - \sum_{k=1}^{N_1} Y_{i,k} Z_{k,i,1}  \right\}
  \\
  &= \sum_{i=1}^N \left\{\log\left(\1^\top_{N_1}e^{Z_{i,1} }\right) 
    - Y^\top_{i} Z_{i,1}  \right\}
  \\
  &= \log(\1_{N_1}^\top e^{Z_1}) \1_N - \tr(Y^\top Z_1)
  \\
  &= \1^\top_N \log\left( \left(I_N \otimes \1^\top_K\right) e^{z_1} \right) - y^\top z_1 
\end{align*}
with $Z_1 \in \Re^{N_1 \times N}$, $N_1 = K$, $y = \vec(Y)$ and $z_1 = \vec(Z_1)$.  
Equation~\eqref{eq:z_L_fb} is now
\begin{align*}
  z_1 & = \argmin_{s}\left\{ \gamma L(y, s) + 
    \frac12 \norm*{s - \eta}^{2}_{\Lambda_{\mu_1}} \right\}\\
\text{where} \; \;  \eta & = P_1 \left( z - \gamma \nabla F(z) \right) \; \text{and} \; \; 
s \in \Re^{N \cdot N_1 \times 1} \;' .
\end{align*}
Since this function has a Lipschitz bounded
derivative with constant $\gamma_1 = 1/4$, we can use 
forward-backward to converge to a solution of the proximal problem.
In this case, the sub-problem forward-backward steps are
\begin{align*}
  s &= \prox_{\norm*{\cdot - \eta}^{2}_{\Lambda_{\mu_1}}/\gamma_1  } 
  \left( s - \frac{\gamma}{\gamma_1} \nabla L(y, s) \right)
  \\
  &= \left(I  + \Lambda_{\mu_1}/\gamma_1 \right)^{-1}\left(
    \frac{\Lambda_{\mu_1}}{\gamma_1} \eta + 
    s - \frac{\gamma}{\gamma_1} \nabla L(y,s)\right)
  \\
  &= \prox_{\Lambda_{\mu_1} L(y, \cdot)} \left( \eta \right) 
\end{align*}
with $p =  e^{z_1} \oslash \left(E_1 e^{z_1} \otimes \1_K\right)$ and
\begin{align*}
  \nabla L(y, z_1) &= p - y  
  \\ 
  \nabla^2 L(y, z_1) &= \Diag(p) - p p^\top
\end{align*}
for $E_1 =  I_N \otimes \1^\top_K$ and $\oslash$ signifying element-wise division. 

All of the methods described here require derivative and/or Hessian information
at some stage in the sub-problem minimization steps, and for the logistic transform
those quantities are easily obtained:
\begin{align*}
 \D f(x) &= 
 \Diag\left(\sigma(x) \odot \left(\1-\sigma(x)\right)\right)
 \\ 
 \H f(x) &=  
 \Diag\left( 
   \sigma(x) \odot \left(\sigma(x) - \1\right) \odot \left(\1 - 2 \sigma(x)\right) 
 \right)
\end{align*}
where $\odot$ is the Hadamard/element-wise product.  In the fully vectorized form
of the model given in Section~\ref{sec:prox_admm}, these quantities will need to
be augmented by the structural properties of mappings like $\tf(z)$.  Such operations
are perhaps best suited for tensor and graph libraries, especially if they're 
capable of distributed the operations and handling the sparsity inherent to 
the model's design.  We do not cover those details here, but our implementations
have shown that standard sparse matrix libraries can be leveraged to
perform such operations without too much coding effort.

\begin{table}[ht]
  \centering
  \begin{tabular}{rlr}
    \hline
    \% non-zero $w$ & $\gamma_w$ & $\mu_l$ \\ 
    \hline
     1.00 & 0 & 0.10 \\ 
     0.54 & 0.67 & 0.10 \\ 
     0.40 & 1.33 & 0.10 \\ 
     0.40 & 2 & 0.10 \\ 
     1.00 & 0 & 1.00 \\ 
     0.40 & 0.67 & 1.00 \\ 
     0.35 & 1.33 & 1.00 \\ 
     0.29 & 2 & 1.00 \\ 
     1.00 & 0 & 1.50 \\ 
     0.33 & 0.67 & 1.50 \\ 
     0.29 & 1.33 & 1.50 \\ 
     0.28 & 2 & 1.50 \\ 
    \hline
  \end{tabular}
  \caption{Percentage of non-zero $w$ entries for the final 
    parameter set in the $\ell^1$ penalized model with logit activations and
    multinomial/softmax loss.}
  \label{tab:w_nonzero}
\end{table}

\subsection{Iris Data}

Fisher's Iris flower dataset consists of four measurements from $50$
observations of three species ($K=3$) of Iris. The species are \emph{Iris setosa, Iris
  virginica} and \emph{Iris versicolor} and we have $150$ total observations.
The various Iris species have petals and sepals (the green sheaths at the base
of the petals) of different lengths which are measured in cm and $ dim(X) = 4 $. The goal is to
predict the label given by species name as a function of the four
characteristics.

To illustrate our methodology, we use a deep learning architecture with a
multinomial (or softmax) loss with a hidden logit layer with a softmax link for $ f_1$. The
hidden layer uses $10$ units and so we have $L=2$ and $N_2 = 10$.
Section~\ref{sec:multinom} provides details of the loss structure and objective
function.  To induce sparsity we use an $\ell^1$ penalization. One purpose of
our study is to show how a sparse model can perform as well as a dense NN
model. The construction of $\phi(\tw)$ includes a sparsity
parameter $\gamma_w$ so that the penalty term is effectively
$$
\phi(\tw) \defeq \gamma_w \sum_{j=1}^{N_{\tw}} |\tw_j|
\;.
$$
By varying $ \gamma_w $ we can find the full regularisation path for our deep learning model much in the same way as a traditional lasso regression.

\begin{figure}[t]
  \begin{center}
  \includegraphics[scale=0.5]{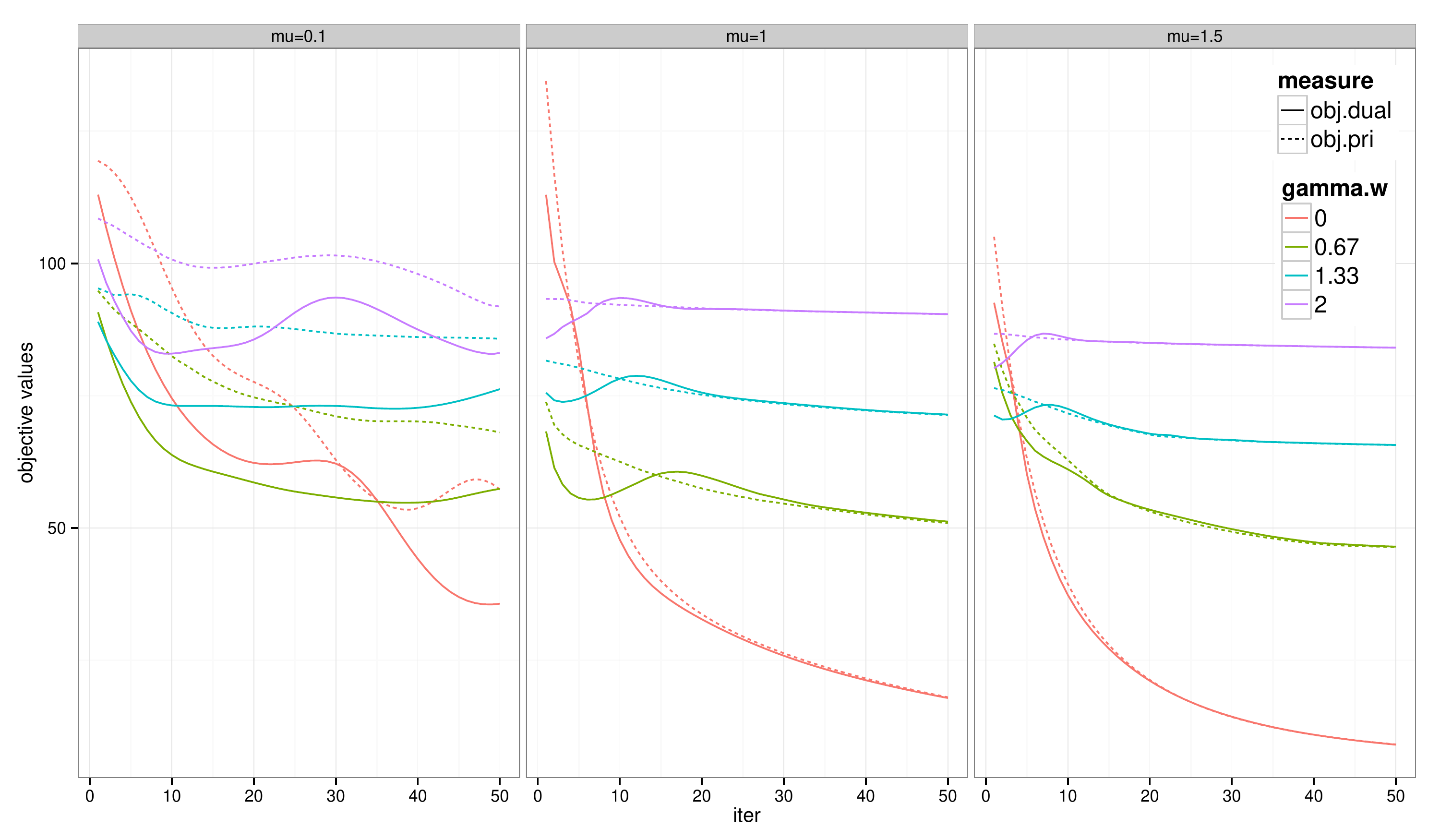}
  \caption{Primal and dual objective values for the $\ell^1$ penalized
    model with logit activations and multinomial/softmax loss.} 
  \label{fig:iris_obj_values}
  \end{center}
\end{figure}

The parameters are estimated from $70$\% of the data, i.e. the training sample,
leaving a hold-out test sample of $30$\%.  
For the same training and testing data,  we find that the neural net library
\texttt{nnet} \cite{venables_modern_2002} in \texttt{R} gives a test error of
$92\%$ on average, which is comparable to the results produced by our model
under the tested configurations.  

\begin{figure}[t]
  \begin{center}
  \includegraphics[scale=0.53]{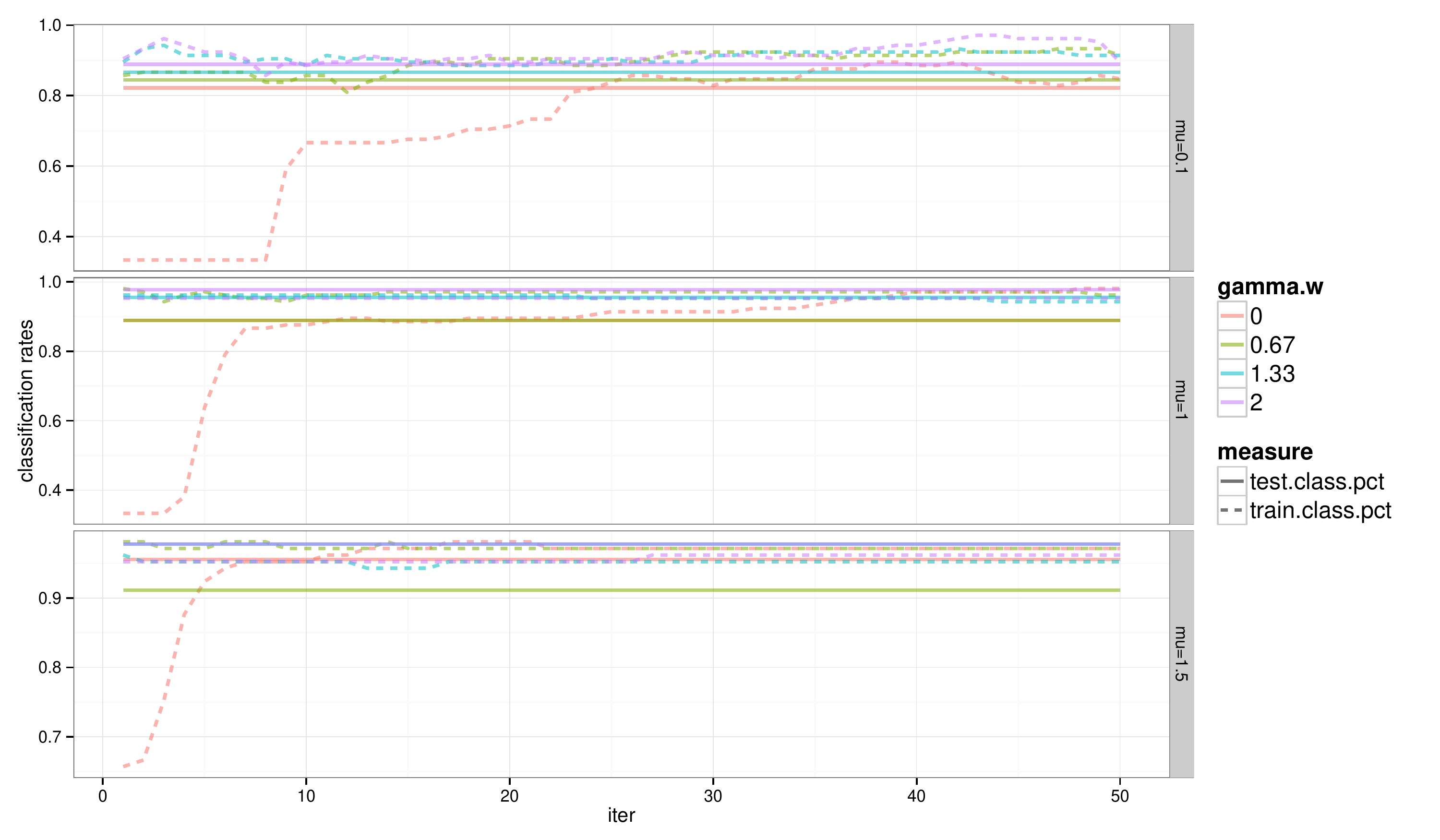}
  \caption{Training and testing classification rates for the $\ell^1$ penalized
    model with logit activations and multinomial/softmax loss.}
  \label{fig:iris_class_values}
  \end{center}
\end{figure}

Figure~\ref{fig:iris_obj_values}
shows the primal and dual objective values across iterations and parameter values.
Figure~\ref{fig:iris_class_values} shows the classification rates.
Table~\ref{tab:w_nonzero} lists the percentage of non-zero entries in $w$
at the end of each fit across the given parameters.  From the plot we can see that
the $\ell^1$ penalization has a noticeable effect over the given range of $\gamma_w$,
and referring back to Figure~\ref{fig:iris_class_values} we see that 
classification rates comparable to the dense $w$ case (i.e. $\gamma_w = 0$)
are obtained for sparse $w$.

\section{Discussion}
\label{sec:discussion}

Deep Learning provides an exciting new area for nonlinear prediction rules in
many applications in image processing and machine learning. High dimensional
data reduction is achieved using a set of hidden layer variables.  Our
estimation methodology uses a proximal splitting techniques that leads to
efficient implementation. Our methods apply to non-convex non-differentiable
regularisation penalties.  The full regularisation path is available as is
hyper-parameter selection via a statistical predictive mean squared error
criterion.

There are many areas of future application. First, there's a range of models
such as convolution neural nets where tailored ADMM methods can be constructed.
Adding acceleration methods such as \cite{nesterov_method_1983} to proximal
operator steps and understanding the speed of convergence of these algorithms
provides another ares of fruitful future research.  There are also alternative
methods for handling the general constrained problem in
\eqref{eq:aug_lagrange}.  The augmented Lagrangian can be defined in terms of a
general penalty instead of the customary $\ell^2$ term.  It would be worthwhile to
investigate which other penalties work well, if not better, for the problems
discussed here.  As well, there is a possible relation between the dropout
technique in Deep Learning and certain types of regularization on $W_l$ and/or
$z_l$.  For that matter, regularization on $z_l$ hasn't been explored in this
context, yet it may provide an automatic means of exploring the space of splitting
designs.

One fruitful area of research is to develop tailored Markov chain Monte Carlo
(MCMC) methods to allow one to correctly assess uncertainty bounds and provide
more efficient predictive rules. For example, a multi-class logistic regression
can be implemented in MCMC using a Polya-Gamma data augmentation
\cite{polson_bayesian_2013} scheme. Another advantage of our regularisation
approach framework is the application to optimal selection of hyper-parameters
defined as the overall amount of regularisation applied to the parameter
weights. The methods in \cite{pereyra_proximal_2013} can be used to combine
proximal approach with MCMC to obtain a full description of the uncertainty in
parameter estimation of the weights.

There are many other areas of future research. For example, tailoring
algorithms for other classes of neural net models such as recurrent neural
networks, see \cite{graves_supervised_2012} is as interesting area of study.
There is also a large literature in statistics on non-parametrically estimating
the link functions for shallow architecture models. Until now, deep learning
pre-specifies the link and it's an open problem to see whether one can
consistently estimate this in a deep layer model.

\newpage
\bibliographystyle{plainnat}
\bibliography{deep-learning}

\begin{appendix}

\section{Block \texorpdfstring{$W_l$}{W l} Steps}
\label{app:w_l_steps}

In this section we expand the quadratic term in \eqref{eq:W_l_prob}.
Using the indexed form of \eqref{eq:aug_lagrange} and
$$
f_{i,l} \defeq f_l(z_{i,l+1}), \quad
W_l f_{i,l} = \left(f_{i,l}^\top \otimes I_{N_l}\right) \vec(W_l), \quad
w_l \defeq \vec(W_l)
$$
we obtain
\begin{align*}
  \sum_{i=1}^{N} \frac{\mu_{i,l}}{2} \norm*{z_{i,l} - W_l f_l(z_{i,l+1}) -b_l+u_{i,l}}^2
  &=
  \sum_{i=1}^{N} \frac{\mu_{i,l}}{2} \norm*{
    \left(f_{i,l}^\top \otimes I_{N_l}\right) w_l - \left(z_{i,l}+u_{i,l}-b_l\right)}^2
\end{align*}
Letting $\xi_{i,l} = z_{i,l}+u_{i,l}-b_l$ and expanding, we get
\begin{gather*}
  \frac12 w_l^\top \Lambda_{w_l} w_l - w_l^\top d_{w_l} + c_{w_l}
\end{gather*}
for
\begin{align*}
  \Lambda_{w_l} &= \sum_{i=1}^{N} \mu_{i,l} f_{i,l} f_{i,l}^\top \otimes I_{N_l}
  = f_l(Z_{l+1}) \Diag(\mu_l) f_l(Z_{l+1})^\top \otimes I_{N_l}
  \\
  d_{w_l} &= \sum_{i=1}^{N} \left(\mu_{i,l} f_{i,l} \otimes I_{N_l}\right) \xi_{i,l}
  = \left(f_l(Z_{l+1}) \Diag(\mu_l) \otimes I_{N_l}\right) \xi_{l}  
  \\ 
  &=\vec\left(\left(Z_l + U_L\right) \Diag(\mu_l) f_l(Z^\top_{l+1})\right) 
          - f_l(Z_{l+1}) \mu_l \otimes b_l
  \\
  c_{w_l} &= \frac12 \sum_{i=1}^{N}\mu_{i,l}\norm*{\xi_{i,l}}^2 
  = \frac12 \norm*{\xi_{l}}^2_{\Lambda_{\mu_l}}
\end{align*}
where $\xi_l = z_l + u_l - \left(\1_{N} \otimes b_l\right)$.
Similarly, we can decompose these terms:
\begin{gather*}
  \Lambda_{w_l} = H_{w_l}^\top H_{w_l} \otimes I_{N_l} 
  = \left(H_{w_l} \otimes I_{N_l} \right)^\top \left(H_{w_l} \otimes I_{N_l}\right) 
  \\
  d_{w_l} = H_{w_l}^\top M_l \otimes I_{N_l} 
  = \left(H_{w_l} \otimes I_{N_l} \right)^\top \left(M_l \otimes I_{N_l}\right) 
\end{gather*}
where $M_l \defeq \Diag(\sqrt{\mu_l}) \in \R^{N \times N}$ and 
$H_{w_l} \defeq M_l f_l(Z^\top_{l+1})  \in \R^{N_{l+1}}$.

\end{appendix}

\end{document}

%% file: math-commands.tex
\newtheorem{Exa}{Example}

\let\D\relax
\DeclareMathOperator*{\D}{D}
\let\H\relax
\DeclareMathOperator*{\H}{H}
\DeclareMathOperator*{\prox}{prox}
\newcommand*{\moreau}[2]{
  \ensuremath{
    {#2}^{
    \ifx&#1&\empty 1 \else #1 \fi
    }
  }
}

\newcommand{\var}{{\rm var}}

\DeclareMathOperator{\Diag}{Diag}

\DeclareMathOperator{\tr}{tr}
\renewcommand{\vec}{\operatorname{vec}}

\newcommand{\R}{{\bf R}}

\renewcommand{\Re}{{\mathbb{R}}}

\newcommand{\1}{{\bf 1}}
\newcommand{\0}{{\bf 0}}

\newcommand{\ben}{\begin{enumerate}}
\newcommand{\een}{\end{enumerate}}
\newcommand{\beq}{\begin{equation}}
\newcommand{\eeq}{\end{equation}}

\DeclareMathOperator*{\argmin}{argmin}
\DeclareMathOperator*{\argmax}{argmax}

\DeclarePairedDelimiter{\norm}{\lVert}{\rVert}

\newcommand{\cX}{\mathcal{X}}
\newcommand{\cZ}{\mathcal{Z}}
\newcommand{\cL}{\mathcal{L}}
\newcommand{\cU}{\mathcal{U}}
\newcommand{\cW}{\mathcal{W}}
\newcommand{\cK}{\mathcal{K}}

\newcommand{\tN}{\tilde{N}}
\newcommand{\tW}{\tilde{W}}
\newcommand{\tw}{\tilde{w}}
\newcommand{\tf}{\tilde{f}}

\newcommand{\defeq}{\coloneqq}


%% file: deep-learning.bbl
\begin{thebibliography}{51}
\providecommand{\natexlab}[1]{#1}
\providecommand{\url}[1]{\texttt{#1}}
\expandafter\ifx\csname urlstyle\endcsname\relax
  \providecommand{\doi}[1]{doi: #1}\else
  \providecommand{\doi}{doi: \begingroup \urlstyle{rm}\Url}\fi

\bibitem[Argyriou et~al.(2013)Argyriou, Cl{\'e}men{\c c}on, and
  Zhang]{argyriou_learning_2013}
Andreas Argyriou, St{\'e}phan Cl{\'e}men{\c c}on, and Ruocong Zhang.
\newblock Learning the graph of relations among multiple tasks.
\newblock 2013.
\newblock URL \url{https://hal.inria.fr/hal-00940321/}.

\bibitem[Bach(2014)]{bach_breaking_2014-1}
Francis Bach.
\newblock Breaking the curse of dimensionality with convex neural networks,
  2014.

\bibitem[Barron(1993)]{barron_universal_1993}
Andrew~R. Barron.
\newblock Universal approximation bounds for superpositions of a sigmoidal
  function.
\newblock \emph{Information Theory, {IEEE} Transactions on}, 39\penalty0
  (3):\penalty0 930--945, 1993.
\newblock URL
  \url{http://ieeexplore.ieee.org/xpls/abs_all.jsp?arnumber=256500}.

\bibitem[Bertsekas(1976)]{bertsekas_multiplier_1976}
Dimitri~P. Bertsekas.
\newblock Multiplier methods: a survey.
\newblock \emph{Automatica}, 12\penalty0 (2):\penalty0 133--145, 1976.
\newblock URL
  \url{http://www.sciencedirect.com/science/article/pii/0005109876900777}.

\bibitem[Bouchard(2007)]{bouchard_efficient_2007}
Guillaume Bouchard.
\newblock Efficient bounds for the softmax function and applications to
  approximate inference in hybrid models.
\newblock pages 1--9, 2007.
\newblock URL \url{http://eprints.pascal-network.org/archive/00003498/}.

\bibitem[Boyd and Vandenberghe(2009)]{boyd_convex_2009}
Stephen Boyd and Lieven Vandenberghe.
\newblock \emph{Convex optimization}.
\newblock Cambridge university press, 2009.

\bibitem[Boyd et~al.(2011)Boyd, Parikh, Chu, Peleato, and
  Eckstein]{boyd_distributed_2011}
Stephen Boyd, Neal Parikh, Eric Chu, Borja Peleato, and Jonathan Eckstein.
\newblock Distributed optimization and statistical learning via the alternating
  direction method of multipliers.
\newblock \emph{Foundations and Trends{\textregistered} in Machine Learning},
  3\penalty0 (1):\penalty0 1--122, 2011.
\newblock URL \url{http://dl.acm.org/citation.cfm?id=2185816}.

\bibitem[Breiman(2001)]{breiman_statistical_2001}
Leo Breiman.
\newblock Statistical modeling: The two cultures (with comments and a rejoinder
  by the author).
\newblock \emph{Statistical Science}, 16\penalty0 (3):\penalty0 199--231, 2001.
\newblock URL \url{http://projecteuclid.org/euclid.ss/1009213726}.

\bibitem[Chen et~al.(2013)Chen, Huang, and Zhang]{chen_primaldual_2013}
Peijun Chen, Jianguo Huang, and Xiaoqun Zhang.
\newblock A primal{\textendash}dual fixed point algorithm for convex separable
  minimization with applications to image restoration.
\newblock \emph{Inverse Problems}, 29\penalty0 (2):\penalty0 025011, 2013.
\newblock ISSN 0266-5611.
\newblock \doi{10.1088/0266-5611/29/2/025011}.
\newblock URL
  \url{http://stacks.iop.org/0266-5611/29/i=2/a=025011?key=crossref.c7722e5930429604a926fdc1944123e0}.

\bibitem[Combettes and Pesquet(2011)]{combettes_proximal_2011}
Patrick~L Combettes and Jean-Christophe Pesquet.
\newblock Proximal splitting methods in signal processing.
\newblock \emph{Fixed-point algorithms for inverse problems in science and
  engineering}, pages 185--212, 2011.

\bibitem[Cook(2007)]{cook_fisher_2007}
R.~Dennis Cook.
\newblock Fisher lecture: Dimension reduction in regression.
\newblock \emph{Statistical Science}, pages 1--26, 2007.
\newblock URL \url{http://www.jstor.org/stable/27645799}.

\bibitem[Cook and Lee(1999)]{cook_dimension_1999}
R.~Dennis Cook and Hakbae Lee.
\newblock Dimension reduction in binary response regression.
\newblock \emph{Journal of the American Statistical Association}, 94\penalty0
  (448):\penalty0 1187--1200, 1999.
\newblock URL
  \url{http://amstat.tandfonline.com/doi/abs/10.1080/01621459.1999.10473873}.

\bibitem[Davison(2003)]{davison_statistical_2003}
Anthony~Christopher Davison.
\newblock \emph{Statistical models}, volume~11.
\newblock Cambridge University Press, 2003.
\newblock URL
  \url{https://books.google.com/books?hl=en\&lr=\&id=gQyIGGAiN4AC\&oi=fnd\&pg=PR9\&dq=davison+statistical+models\&ots=0UbFWpefvz\&sig=-NXdEmD8vBq9ct4oF74R6NeW4Jk}.

\bibitem[Dean et~al.(2012)Dean, Corrado, Monga, Chen, Devin, Mao, Senior,
  Tucker, Yang, Le, and {others}]{dean_large_2012}
Jeffrey Dean, Greg Corrado, Rajat Monga, Kai Chen, Matthieu Devin, Mark Mao,
  Andrew Senior, Paul Tucker, Ke~Yang, Quoc~V. Le, and {others}.
\newblock Large scale distributed deep networks.
\newblock In \emph{Advances in Neural Information Processing Systems}, pages
  1223--1231, 2012.
\newblock URL
  \url{http://papers.nips.cc/paper/4687-large-scale-distributed-deep-networks}.

\bibitem[Dellaportas and Smith(1993)]{dellaportas_bayesian_1993}
P.~Dellaportas and Adrian~{FM} Smith.
\newblock Bayesian inference for generalized linear and proportional hazards
  models via gibbs sampling.
\newblock \emph{Applied Statistics}, pages 443--459, 1993.
\newblock URL \url{http://www.jstor.org/stable/2986324}.

\bibitem[Efron(1983)]{efron_estimating_1983}
Bradley Efron.
\newblock Estimating the error rate of a prediction rule: Improvement on
  cross-validation.
\newblock \emph{Journal of the American Statistical Association}, 78\penalty0
  (382):\penalty0 316--331, 1983.

\bibitem[Farabet et~al.(2013)Farabet, Couprie, Najman, and
  {LeCun}]{farabet_learning_2013}
Clement Farabet, Camille Couprie, Laurent Najman, and Yann {LeCun}.
\newblock Learning hierarchical features for scene labeling.
\newblock \emph{Pattern Analysis and Machine Intelligence, {IEEE} Transactions
  on}, 35\penalty0 (8):\penalty0 1915--1929, 2013.
\newblock URL
  \url{http://ieeexplore.ieee.org/xpls/abs_all.jsp?arnumber=6338939}.

\bibitem[Genkin et~al.(2007)Genkin, Lewis, and Madigan]{genkin_largescale_2007}
Alexander Genkin, David~D. Lewis, and David Madigan.
\newblock Large-scale bayesian logistic regression for text categorization.
\newblock \emph{Technometrics}, 49\penalty0 (3):\penalty0 291--304, 2007.
\newblock URL
  \url{http://amstat.tandfonline.com/doi/abs/10.1198/004017007000000245}.

\bibitem[George and Foster(2000)]{george_calibration_2000}
{EdwardI} George and Dean~P. Foster.
\newblock Calibration and empirical bayes variable selection.
\newblock \emph{Biometrika}, 87\penalty0 (4):\penalty0 731--747, 2000.
\newblock URL \url{http://biomet.oxfordjournals.org/content/87/4/731.short}.

\bibitem[Graves and {others}(2012)]{graves_supervised_2012}
Alex Graves and {others}.
\newblock \emph{Supervised sequence labelling with recurrent neural networks},
  volume 385.
\newblock Springer, 2012.
\newblock URL
  \url{http://link.springer.com/content/pdf/10.1007/978-3-642-24797-2.pdf}.

\bibitem[Green et~al.(2015)Green, \ensuremath{\backslash}Latuszy{\'n}ski,
  Pereyra, and Robert]{green_bayesian_2015}
P.~J. Green, K.~\ensuremath{\backslash}Latuszy{\'n}ski, M.~Pereyra, and C.~P.
  Robert.
\newblock Bayesian computation: a perspective on the current state, and
  sampling backwards and forwards.
\newblock \emph{{ArXiv} e-prints}, February 2015.

\bibitem[Hinton and Salakhutdinov(2006)]{hinton_reducing_2006}
Geoffrey~E. Hinton and Ruslan~R. Salakhutdinov.
\newblock Reducing the dimensionality of data with neural networks.
\newblock \emph{Science}, 313\penalty0 (5786):\penalty0 504--507, 2006.
\newblock URL \url{http://www.sciencemag.org/content/313/5786/504.short}.

\bibitem[Hinton et~al.(2006)Hinton, Osindero, and Teh]{hinton_fast_2006}
Geoffrey~E. Hinton, Simon Osindero, and Yee-Whye Teh.
\newblock A fast learning algorithm for deep belief nets.
\newblock \emph{Neural computation}, 18\penalty0 (7):\penalty0 1527--1554,
  2006.
\newblock URL
  \url{http://www.mitpressjournals.org/doi/abs/10.1162/neco.2006.18.7.1527}.

\bibitem[Hurvich and Tsai(1991)]{hurvich_bias_1991}
Clifford~M. Hurvich and Chih-Ling Tsai.
\newblock Bias of the corrected {AIC} criterion for underfitted regression and
  time series models.
\newblock \emph{Biometrika}, 78\penalty0 (3):\penalty0 499--509, 1991.
\newblock URL \url{http://biomet.oxfordjournals.org/content/78/3/499.short}.

\bibitem[Knowles and Minka(2011)]{knowles_nonconjugate_2011}
David~A. Knowles and Tom Minka.
\newblock Non-conjugate variational message passing for multinomial and binary
  regression.
\newblock In \emph{Advances in Neural Information Processing Systems}, pages
  1701--1709, 2011.
\newblock URL
  \url{http://papers.nips.cc/paper/4407-non-conjugate-variational-message-passing-for-multinomial-and-binary-regression}.

\bibitem[Kolmogorov(1957)]{kolmogorov_representation_1957}
Andre{\u \i} Kolmogorov.
\newblock The representation of continuous functions of many variables by
  superposition of continuous functions of one variable and addition.
\newblock \emph{Dokl. Akad. Nauk {SSSR}}, 114:\penalty0 953--956, 1957.

\bibitem[{LeCun} et~al.(2012){LeCun}, Bottou, Orr, and
  M{\"u}ller]{lecun_efficient_2012}
Yann~A. {LeCun}, L{\'e}on Bottou, Genevieve~B. Orr, and Klaus-Robert
  M{\"u}ller.
\newblock Efficient backprop.
\newblock In \emph{Neural networks: Tricks of the trade}, pages 9--48.
  Springer, 2012.
\newblock URL
  \url{http://link.springer.com/chapter/10.1007/978-3-642-35289-8_3}.

\bibitem[Lewis and Wright(2008)]{lewis_proximal_2008}
Adrian~S. Lewis and Stephen~J. Wright.
\newblock A proximal method for composite minimization.
\newblock \emph{{arXiv} preprint {arXiv}:0812.0423}, 2008.
\newblock URL \url{http://arxiv.org/abs/0812.0423}.

\bibitem[Lorenz(1963)]{lorenz_deterministic_1963}
Edward~N. Lorenz.
\newblock Deterministic nonperiodic flow.
\newblock \emph{Journal of the atmospheric sciences}, 20\penalty0 (2):\penalty0
  130--141, 1963.
\newblock URL
  \url{http://journals.ametsoc.org/doi/abs/10.1175/1520-0469(1963)020\%3C0130:DNF\%3E2.0.CO;2}.

\bibitem[Madigan et~al.(2005)Madigan, Genkin, Lewis, Fradkin, and
  {others}]{madigan_bayesian_2005}
David Madigan, Alexander Genkin, David~D. Lewis, Dmitriy Fradkin, and {others}.
\newblock Bayesian multinomial logistic regression for author identification.
\newblock \emph{Bayesian Inference and Maximum Entropy Methods in Science and
  Engineering}, 803:\penalty0 509--516, 2005.
\newblock URL
  \url{http://citeseerx.ist.psu.edu/viewdoc/download?doi=10.1.1.60.2085\&rep=rep1\&type=pdf}.

\bibitem[Martens and Sutskever(2011)]{martens_learning_2011}
James Martens and Ilya Sutskever.
\newblock Learning recurrent neural networks with hessian-free optimization.
\newblock In \emph{Proceedings of the 28th International Conference on Machine
  Learning ({ICML}-11)}, pages 1033--1040, 2011.
\newblock URL
  \url{http://machinelearning.wustl.edu/mlpapers/paper_files/ICML2011Martens_532.pdf}.

\bibitem[Masci et~al.(2013)Masci, Giusti, Ciresan, Fricout, and
  Schmidhuber]{masci_fast_2013}
Jonathan Masci, Alessandro Giusti, Dan Ciresan, Gabriel Fricout, and J{\"u}rgen
  Schmidhuber.
\newblock A fast learning algorithm for image segmentation with max-pooling
  convolutional networks.
\newblock In \emph{Image Processing ({ICIP}), 2013 20th {IEEE} International
  Conference on}, pages 2713--2717. {IEEE}, 2013.
\newblock URL
  \url{http://ieeexplore.ieee.org/xpls/abs_all.jsp?arnumber=6738559}.

\bibitem[Mnih et~al.(2013)Mnih, Kavukcuoglu, Silver, Graves, Antonoglou,
  Wierstra, and Riedmiller]{mnih_playing_2013}
Volodymyr Mnih, Koray Kavukcuoglu, David Silver, Alex Graves, Ioannis
  Antonoglou, Daan Wierstra, and Martin Riedmiller.
\newblock Playing atari with deep reinforcement learning.
\newblock \emph{{arXiv} preprint {arXiv}:1312.5602}, 2013.
\newblock URL \url{http://arxiv.org/abs/1312.5602}.

\bibitem[Nesterov(1983)]{nesterov_method_1983}
Yurii Nesterov.
\newblock A method of solving a convex programming problem with convergence
  rate {\textdollar}o(1/k\^{}2){\textdollar}.
\newblock In \emph{Soviet Mathematics Doklady}, volume~27, pages 372--376,
  1983.

\bibitem[Ngiam et~al.(2011)Ngiam, Coates, Lahiri, Prochnow, Le, and
  Ng]{ngiam_optimization_2011}
Jiquan Ngiam, Adam Coates, Ahbik Lahiri, Bobby Prochnow, Quoc~V. Le, and
  Andrew~Y. Ng.
\newblock On optimization methods for deep learning.
\newblock In \emph{Proceedings of the 28th International Conference on Machine
  Learning ({ICML}-11)}, pages 265--272, 2011.
\newblock URL
  \url{http://machinelearning.wustl.edu/mlpapers/paper_files/ICML2011Le_210.pdf}.

\bibitem[Orr(1995)]{orr_regularization_1995}
Mark~{JL} Orr.
\newblock Regularization in the selection of radial basis function centers.
\newblock \emph{Neural computation}, 7\penalty0 (3):\penalty0 606--623, 1995.
\newblock URL
  \url{http://www.mitpressjournals.org/doi/abs/10.1162/neco.1995.7.3.606}.

\bibitem[Paige and Butler(2001)]{paige_bayesian_2001}
Robert~L. Paige and Ronald~W. Butler.
\newblock Bayesian inference in neural networks.
\newblock \emph{Biometrika}, 88\penalty0 (3):\penalty0 623--641, 2001.
\newblock URL \url{http://biomet.oxfordjournals.org/content/88/3/623.short}.

\bibitem[Parikh and Boyd(2014)]{parikh_proximal_2014}
Neal Parikh and Stephen Boyd.
\newblock Proximal algorithms.
\newblock \emph{Foundations and Trends in Optimization}, 1\penalty0
  (3):\penalty0 123--231, 2014.
\newblock ISSN 2167-3888.
\newblock \doi{10.1561/2400000003}.

\bibitem[Pereyra(2013)]{pereyra_proximal_2013}
Marcelo Pereyra.
\newblock Proximal markov chain monte carlo algorithms.
\newblock \emph{{arXiv} preprint {arXiv}:1306.0187}, 2013.
\newblock URL \url{http://arxiv.org/abs/1306.0187}.

\bibitem[Poggio and Girosi(1990)]{poggio_networks_1990}
Tomaso Poggio and Federico Girosi.
\newblock Networks for approximation and learning.
\newblock \emph{Proceedings of the {IEEE}}, 78\penalty0 (9):\penalty0
  1481--1497, 1990.
\newblock URL \url{http://ieeexplore.ieee.org/xpls/abs_all.jsp?arnumber=58326}.

\bibitem[Polson and Scott(2013)]{polson_data_2013}
Nicholas~G. Polson and James~G. Scott.
\newblock Data augmentation for non-gaussian regression models using
  variance-mean mixtures.
\newblock \emph{Biometrika}, 100\penalty0 (2):\penalty0 459--471, 2013.
\newblock ISSN 00063444.
\newblock \doi{10.1093/biomet/ass081}.

\bibitem[Polson et~al.(2011)Polson, Scott, and {others}]{polson_data_2011}
Nicholas~G. Polson, Steven~L. Scott, and {others}.
\newblock Data augmentation for support vector machines.
\newblock \emph{Bayesian Analysis}, 6\penalty0 (1):\penalty0 1--23, 2011.
\newblock URL \url{http://projecteuclid.org/euclid.ba/1339611936}.

\bibitem[Polson et~al.(2013)Polson, Scott, and Windle]{polson_bayesian_2013}
Nicholas~G. Polson, James~G. Scott, and Jesse Windle.
\newblock Bayesian inference for logistic models using
  p{\'o}lya{\textendash}gamma latent variables.
\newblock \emph{Journal of the American Statistical Association}, 108\penalty0
  (504):\penalty0 1339--1349, 2013.
\newblock URL
  \url{http://www.tandfonline.com/doi/abs/10.1080/01621459.2013.829001}.

\bibitem[Polson et~al.(2015)Polson, Scott, and Willard]{polson_proximal_2015}
Nicholas~G. Polson, James~G. Scott, and Brandon~T. Willard.
\newblock Proximal algorithms in statistics and machine learning.
\newblock pages 1--43, 2015.

\bibitem[Ripley(1996)]{ripley_pattern_1996}
Brian~D. Ripley.
\newblock \emph{Pattern recognition and neural networks}.
\newblock Cambridge university press, 1996.
\newblock URL
  \url{https://books.google.com/books?hl=en\&lr=\&id=2SzT2p8vP1oC\&oi=fnd\&pg=PR9\&dq=ripley+neural+network\&ots=JLTUg5dRXL\&sig=mTMtffAcI4p1lpotV60k6c3vMWk}.

\bibitem[Stein(1981)]{stein_estimation_1981}
Charles~M. Stein.
\newblock Estimation of the mean of a multivariate normal distribution.
\newblock \emph{The Annals of Statistics}, 9\penalty0 (6):\penalty0 1135--1151,
  1981.
\newblock ISSN 0090-5364.
\newblock \doi{10.1214/aos/1176345632}.

\bibitem[Titterington(2004)]{titterington_bayesian_2004}
D.~M. Titterington.
\newblock Bayesian methods for neural networks and related models.
\newblock \emph{Statistical Science}, 19\penalty0 (1):\penalty0 128--139, 2004.

\bibitem[Venables and Ripley(2002)]{venables_modern_2002}
W.~N. Venables and B.~D. Ripley.
\newblock \emph{Modern Applied Statistics with S}.
\newblock Springer, New York, fourth edition, 2002.
\newblock URL \url{http://www.stats.ox.ac.uk/pub/MASS4}.
\newblock ISBN 0-387-95457-0.

\bibitem[Wang and Carreira-Perpin{\'a}n(2012)]{wang_nonlinear_2012}
Weiran Wang and Miguel~A. Carreira-Perpin{\'a}n.
\newblock Nonlinear low-dimensional regression using auxiliary coordinates.
\newblock In \emph{International Conference on Artificial Intelligence and
  Statistics}, pages 1295--1304, 2012.
\newblock URL
  \url{http://machinelearning.wustl.edu/mlpapers/paper_files/AISTATS2012_WangC12.pdf}.

\bibitem[Windle et~al.(2013)Windle, Carvalho, Scott, and
  Sun]{windle_efficient_2013}
Jesse Windle, Carlos~M. Carvalho, James~G. Scott, and Liang Sun.
\newblock Efficient data augmentation in dynamic models for binary and count
  data.
\newblock \emph{{arXiv} preprint {arXiv}:1308.0774}, 2013.
\newblock URL \url{http://arxiv.org/abs/1308.0774}.

\bibitem[Wold(1956)]{wold_causal_1956}
Herman Wold.
\newblock Causal inference from observational data: A review of end and means.
\newblock \emph{Journal of the Royal Statistical Society. Series A (General)},
  pages 28--61, 1956.
\newblock URL \url{http://www.jstor.org/stable/2342961}.

\end{thebibliography}
